\documentclass{article}

\usepackage{arxiv}

\usepackage[utf8]{inputenc} % allow utf-8 input
\usepackage[T1]{fontenc}    % use 8-bit T1 fonts
\usepackage{hyperref}       % hyperlinks
\usepackage{url}            % simple URL typesetting
\usepackage{booktabs}       % professional-quality tables
\usepackage{amsfonts}       % blackboard math symbols
\usepackage{nicefrac}       % compact symbols for 1/2, etc.
\usepackage{microtype}      % microtypography
\usepackage{graphicx}
\usepackage{float}
\usepackage{longtable}
\usepackage{natbib}
\usepackage{doi}

\title{Benchmarking Vision-Language Models for French PDF-to-Markdown Conversion}

\author{\hspace{1mm}Bruno Rigal \\
	Probayes, La Poste \\
	\texttt{bruno.rigal@probayes.com} \\
	\And
	\hspace{1mm}Victor Dupriez \\
	Probayes, La Poste \\
	\texttt{victor.dupriez@probayes.com} \\
	\And
	\hspace{1mm}Alexis Mignon \\
	Probayes, La Poste \\
	\texttt{alexis.mignon@probayes.com}
	\And
	\hspace{1mm}Ronan Le Hy \\
	Probayes, La Poste \\
	\texttt{ronan.lehy@probayes.com}
	\And
	\hspace{1mm}Nicolas Mery \\
	OpenValue, La Poste \\
	\texttt{nicolas.mery@openvalue.com}
}
\renewcommand{\shorttitle}{French PDF-to-Markdown Benchmark}

\hypersetup{
hypertexnames=false,
pdftitle={Benchmarking Vision-Language Models for French PDF-to-Markdown Conversion},
pdfsubject={Document AI, OCR, PDF parsing, Vision-Language Models},
pdfauthor={Brigal},
pdfkeywords={OCR, PDF-to-Markdown, benchmark, French, vision-language models},
}

\begin{document}
\maketitle

\begin{abstract}
This report evaluates PDF-to-Markdown conversion using recent Vision-Language Models (VLMs) on challenging French documents. Document parsing is a critical step for Retrieval-Augmented Generation (RAG) pipelines, where transcription and layout errors propagate to downstream retrieval and grounding. Existing benchmarks often emphasize English or Chinese and can over-penalize benign formatting and linearization choices (e.g., line breaks, list segmentation, alternative table renderings) that are largely irrelevant for downstream use.

We introduce a French-focused benchmark of difficult pages selected via model-disagreement sampling from a corpus of 60{,}000 documents, covering handwritten forms, complex layouts, dense tables, and graphics-rich pages. Evaluation is performed with unit-test-style checks that target concrete failure modes (text presence, reading order, and local table constraints) combined with category-specific normalization designed to discount presentation-only variance. Across 15 models, we observe substantially higher robustness for the strongest proprietary models on handwriting and forms, while several open-weights systems remain competitive on standard printed layouts.
\end{abstract}

% keywords can be removed

\section{Introduction}
The conversion of unstructured PDF documents into structured text formats such as Markdown is a critical step for modern document pipelines, especially for Retrieval-Augmented Generation (RAG). Errors introduced at this stage propagate directly to downstream retrieval, summarization, and grounding.

Recent surveys on visually rich document understanding \citep{ding_survey_2025} emphasize that, despite rapid progress in vision-language models, evaluation remains fragmented and task-dependent. A broad set of benchmarks has emerged for document VQA and extraction (e.g., DocVQA) and for OCR robustness (e.g., OCRBench) \citep{liu_ocrbench_2024,fu_ocrbench_2025}. More recent datasets expand the diversity of layouts and modalities: MosaicDoc stresses complex magazine/news pages \citep{chen_mosaicdoc_2025}, UniDoc-Bench targets multimodal RAG behaviors \citep{peng_unidoc-bench_2026}, and handwriting-focused resources such as HW-MLVQA and NoTeS-Bank probe handwritten content and grounding \citep{pal_hw-mlvqa_2025,pal_notes-bank_2025}.

Two benchmarks are particularly relevant to PDF-to-Markdown conversion. OmniDocBench \citep{ouyang_omnidocbench_2025} provides Chinese--English pages with fine-grained annotations (text, tables, formulas, layout, reading order) and a unified scoring protocol combining edit distance, TEDS, detection metrics, and formula-structure similarity. It is an excellent stress test for end-to-end parsing, but it is restricted to two languages and largely to clean or semi-clean scans.

At the same time, our setting exposes limitations of common scoring choices. Global edit-distance metrics on the whole page (e.g., character error rate or sequence-level Levenshtein distance) become less relevant once transcription is mostly correct: residual differences are often dominated by arbitrary linearization and formatting decisions rather than recognition errors. This brittleness is particularly acute for PDF-to-Markdown, where multiple outputs can be semantically equivalent (e.g., different line-break policies, list vs. paragraph segmentation, caption placement, or alternative table renderings) yet still produce large string distances. For multi-column pages, marginalia, and figure-heavy layouts, ``the'' ground-truth reading order is also intrinsically ambiguous; acceptable region traversals can differ while preserving content, which further inflates edit-distance scores. As a result, global string metrics can over-penalize harmless structural variance while under-emphasizing the operational failure modes that matter in practice (hallucinated spans, OCR substitutions, dropped entities, and repetition/looping behaviors).

The OlmOCR benchmark takes a complementary approach by evaluating document-level OCR through collections of small, machine-checkable ``facts'' about each page \citep{poznanski_OlmOCR_2025,poznanski_OlmOCR_2025-1}. This unit-test framing avoids a single global distance and is effective for diagnosing structural errors and subtle hallucinations. However, OlmOCR-bench focuses primarily on English PDFs and can still be sensitive to benign formatting and normalization choices (e.g., whitespace, punctuation, Unicode variants) that are not necessarily harmful for downstream use. Indeed, our manual audit of failure cases (detailed in Supplementary Material) indicates that for strongly performing models, more than 50\% of the detected failures in OlmOCR-bench are attributable to annotation noise or ambiguities rather than genuine transcription errors in our audited sample of 59 failures. 

Finally, many existing datasets under-represent failure modes that are critical for operational document pipelines: tiny printed text on low-quality scans, long multi-column articles, dense tables spanning entire pages, mixed handwritten--printed forms, and graphics-rich layouts. Practitioners therefore lack a benchmark that simultaneously (i) targets French documents, (ii) emphasizes hard real-world layouts and handwriting, and (iii) evaluates outputs in a way aligned with their use as RAG inputs, where semantic completeness and structural coherence in Markdown often matter more than exact string fidelity.

\begin{table}[htbp]
    \centering
    \caption{Comparison of OCR benchmarks}
    \label{tab:ocr_benchmarks}
    \renewcommand{\arraystretch}{1.3} % Slightly increases row spacing
	\setlength{\tabcolsep}{3pt}
	\resizebox{\linewidth}{!}{%
	\begin{tabular}{l p{1.8cm} p{2.5cm} p{4cm} p{3.5cm}}
        \toprule
        \textbf{Benchmark} & \textbf{Size} & \textbf{Languages} & \textbf{Document Types} & \textbf{Metrics} \\
        \midrule
        \textbf{OCRBench v2} & 850 pages (subset) & English, Chinese & Slides, Papers, Web Documents & Edit Distance (Levenshtein) \\
        \midrule
        \textbf{CC-OCR} & 300 pages (subset) & Multilingual (10 incl. FR, JP, KR) & Magazines, Brochures, Creative Webpages & Edit Distance / BLEU \\
        \midrule
        \textbf{OmniDocBench 1.5} & 1355 pages & English, Chinese & Academic, Manuals, Handwritten Notes & Hierarchical: Text, TEDS (Tables), CDM (Formulas) \\
        \midrule
        \textbf{OlmOCR Bench} & 1403 pages & English, Chinese & Legal PDFs, Academic, Manuals & Unit Tests (Semantic Verification) \\
        \bottomrule
    \end{tabular}
	}
\end{table}

This report addresses these gaps by introducing a specialized benchmark to assess VLM performance on challenging French document layouts for PDF-to-Markdown conversion. We focus on practical, diagnosable failure modes using unit-test-style evaluations with category-specific normalization, prioritizing semantic completeness, structural coherence, and robustness of the produced Markdown for downstream LLM pipelines.

\paragraph{Contributions.}
We make three contributions: (i) a French-focused benchmark of difficult PDF pages selected via model-disagreement sampling, spanning handwriting, forms, complex layouts, and dense tables; (ii) a unit-test-based evaluation suite for PDF-to-Markdown with category-specific normalization profiles designed to discount presentation-only variance; and (iii) a comparative study of 15 proprietary and open-weights models under a unified conversion pipeline.

\section{Methodology}
\subsection{Task definition}
The primary task is the conversion of PDF page images into Markdown text. This requires Optical Character Recognition (OCR), layout analysis (identifying headers, lists, and reading order), and table structure recognition. Models must also interpret non-textual elements, such as converting handwritten fields in forms or describing scientific graphics.

\subsection{Dataset construction}
The source corpus consists of approximately 60,000 French documents from CCPDF and Gallica. To ensure the benchmark targets processing limitations rather than trivial cases, we employed an adversarial selection strategy. Documents were transcribed by two different VLMs (dots-ocr and mineru2.5), and we used the edit distance between their outputs as a proxy for model disagreement. Pages with the largest disagreements were selected as candidates for ``difficulty,'' which intentionally biases the benchmark toward pages where current conversion systems are unstable (rather than implying an absolute, ground-truth notion of difficulty).

The term ``difficulty'' here refers to pages where models disagree strongly under the same conversion setting, which increases the prevalence of operational failure modes (e.g., omissions, substitutions, reading-order mistakes, and structural degradation) compared to random sampling.

The resulting dataset is categorized into specific challenge types:
\begin{itemize}
\item Long tiny text: Dense pages requiring sustained generation window attention with low-resolution or small font sizes.
\item Multi-column: Complex flows requiring correct reading order capabilities.
\item Long tables: Data-dense structures spanning large vertical space.
\item Manuscript/handwritten: Historical documents and contemporary handwritten notes.
\item Forms: Semi-structured documents with handwritten entries. These forms were written manually by a diverse set of participants to ensure variability of handwriting styles.
\item Graphics: Scientific figures and charts requiring textual description. Note that many VLMs were not trained explicitly for this task, which may be judged unfair. However we include this category to reflect real-world use cases where graphics often are the most information-dense part of a document. For downstream RAG and analytics pipelines, failing to transcribe such regions is operationally equivalent to dropping content, which further justifies our choice to penalize omissions of charts and figures.
\end{itemize}
Mathematical equations were explicitly excluded, as they are extensively covered in other benchmarks \citep{ouyang_omnidocbench_2025,poznanski_OlmOCR_2025}.

\subsection{Test generation and verification}
Tests were generated semi-automatically using category-specific prompts and verified via a human-in-the-loop process. For each selected page, gemini-3-pro was prompted with tailored instructions (e.g., ``identify the tiniest text spans'', ``list handwritten fields'', ``extract table cells around a given header'') and its textual candidates were converted into machine-checkable unit tests.

To streamline this verification, we built a lightweight Streamlit application that supports rapid review and correction. The interface displays the page image alongside the model transcription and the candidate test span, with a diff-style visualization to highlight mismatches and missing tokens. This setup enables annotators to quickly adjust the target text or normalization settings, reducing turnaround time while improving the consistency of the final unit tests.

This procedure was applied per category:
\begin{itemize}
\item Tiny text, long text, multi-column, and graphics pages primarily yielded TextPresenceTest items that assert that a short span must appear (or, for headers/footers, must be absent) somewhere in the Markdown. The test assert the presence of verbatim text inside figures instead of the more ambiguous and difficult to judge quality of a semantic captioning.
\item Handwritten and forms pages combined TextPresenceTest items with stricter normalization settings to tolerate minor spacing and Unicode differences while still flagging gross transcription errors.
\item Long tables produced TableTest instances that encode local structural constraints derived from manual inspection of the rendered table.
\end{itemize}

\subsection{Test types and normalization}
All unit tests inherit from a common BasePDFTest class, which is responsible for normalizing both the reference text (from the annotation) and the candidate text (from the model output) before comparison. The normalization is designed to reduce spurious failures caused by harmless formatting differences while still being sensitive to genuine content errors, in the spirit of unit-test-based OCR evaluation \citep{poznanski_OlmOCR_2025-1}.

Concretely, normalization proceeds in several stages activated based on per-test requirements:
\begin{itemize}
\item Global markdown/HTML cleanup: flatten line breaks and HTML-style breaks, remove markdown emphasis markers (bold/italics), keep all tables in html and translate markdown tables to html, and collapse whitespace.
\item Unicode harmonization and symbol mapping: normalize to a canonical Unicode form and apply a curated replacement table (quotes, dashes, checkbox symbols, French guillemets).
\item Optional ASCII projection: when enabled, project accented characters to ASCII to reduce brittleness across OCR systems.
\item Alphanumeric filtering: optionally strip most non-alphanumeric characters and lowercase for numeric/ID-heavy tests.
\item Layout-insensitive spacing control: optionally discard intra-line spaces and/or line breaks.
\item Fine-grained masking: optionally remove specific characters or substrings on a per-test basis.
\end{itemize}

By tuning these options per test type and per category, we obtain a spectrum of strictness. Overall, the improved normalization reduces false negatives due to superficial formatting variations while making failures more interpretable.

\section{Evaluation protocol}
Experiments were conducted using the \texttt{vlmparse} library (\url{https://github.com/ld-lab-pulsia/vlmparse}), a unified wrapper for vision-language models and OCR solutions to parse PDF documents into Markdown. \texttt{vlmparse} provides a consistent conversion interface across heterogeneous providers, supports asynchronous/concurrent processing for high throughput, and can automatically manage Docker-based local model servers for open-source converters. It also includes a Streamlit-based viewer to inspect conversion results. In this work, \texttt{vlmparse} was used both to run proprietary models (e.g., Gemini and OpenAI GPT) and to evaluate a range of open-source OCR/VLM converters under a standardized pipeline.

Inference was parallelized across 32 threads on an NVIDIA A100 (80 GB VRAM) GPU with a timeout of 500s. For a faire comparison, only the inference throughput of local models were compared.

We evaluated a mix of proprietary and open-weights models. For the open and research models, we include representative recent OCR/document parsing systems such as OlmOCR \citep{poznanski_OlmOCR_2025,poznanski_OlmOCR_2025-1}, LightOnOCR \citep{taghadouini_lightonocr_2026}, dots.ocr \citep{li_dotsocr_2025}, DeepSeek-OCR \citep{wei_deepseek-ocr_2025}, and PaddleOCR-VL \citep{cui_paddleocr-vl_2025}. The metric was the pass rate of unit tests checking for specific content strings, reading order correctness, and table structure integrity. Throughput is measured in seconds per page.

The primary metric is the unit-test pass rate, aggregating TextPresenceTest, TextOrderTest, and TableTest instances. This metric makes it very clear what are the requirements of an ocr system and allows a fast overview of failure modes of an OCR system through simple diff viewing available in our open-sourced \href{https://github.com/ld-lab-pulsia/benchpdf2md}{benchmark code}. Throughput is measured in seconds per page on our hardware/software stack.

\section{Results}
\subsection{Overall performance}
\begin{figure}[t]
\centering
\includegraphics[width=0.65\linewidth]{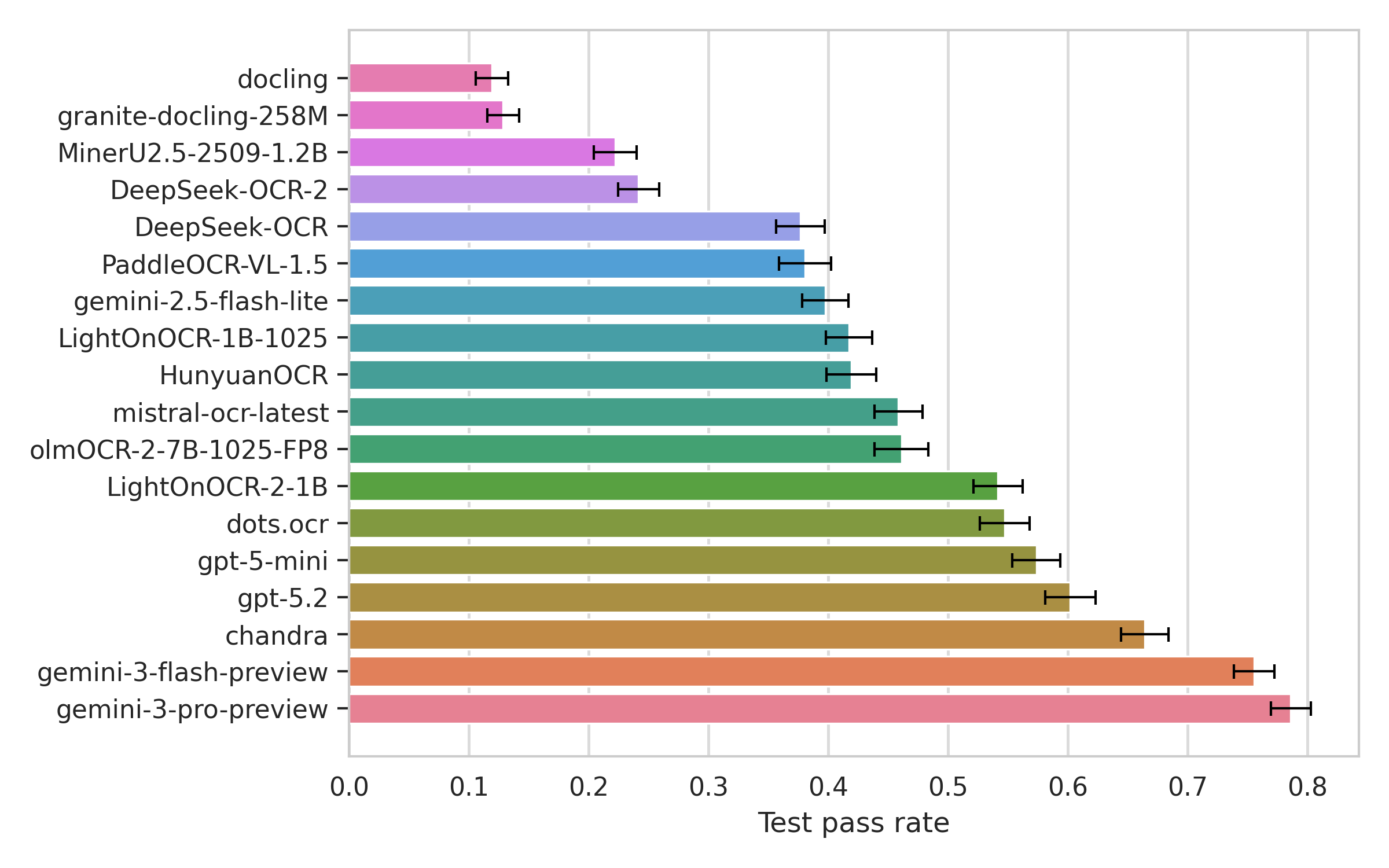}
\caption{Overall benchmark results by model (aggregate unit-test pass rate).}
\label{fig:barplot_results}
\end{figure}

The aggregate performance metrics demonstrate a clear stratification between model classes. The Gemini 3 Pro Preview achieved the highest average score (0.76), followed closely by Gemini 3 Flash Preview (0.74). The highest-performing open-weights model was Chandra (0.66). Figure~\ref{fig:barplot_results} summarizes overall results across models, and Table~\ref{tab:model_category_results} reports per-category performance.

\begin{table}[t]
\caption{Category results by model (mean scores).}
\label{tab:model_category_results}
\centering
\small
\setlength{\tabcolsep}{3pt}
\resizebox{\linewidth}{!}{%
\begin{tabular}{lccccccccc}
\toprule
 & baseline & forms & graphics & handwritten & long\_table & multicolumn & tiny\_text & Time per page [s] & All categories \\
\midrule
gemini-3-pro-preview & $0.965$ & $\textbf{0.725}$ & $0.773$ & $\textbf{0.600}$ & $0.813$ & $\textbf{0.867}$ & $\textbf{0.831}$ &  & $\textbf{0.786}$ \\
gemini-3-flash-preview & $0.964$ & $0.684$ & $0.734$ & $0.582$ & $\textbf{0.828}$ & $\textbf{0.867}$ & $0.804$ &  & $0.755$ \\
datalab-to/chandra & $0.996$ & $0.375$ & $0.748$ & $0.212$ & $0.722$ & $0.794$ & $0.765$ & $4.290$ & $0.664$ \\
gpt-5.2 & $0.998$ & $0.481$ & $0.802$ & $0.206$ & $0.732$ & $0.739$ & $0.535$ &  & $0.602$ \\
gpt-5-mini & $\textbf{1.000}$ & $0.416$ & $\textbf{0.816}$ & $0.182$ & $0.660$ & $0.770$ & $0.506$ &  & $0.574$ \\
rednote-hilab/dots.ocr & $0.988$ & $0.351$ & $0.269$ & $0.079$ & $0.628$ & $0.782$ & $0.765$ & $2.432$ & $0.547$ \\
lightonai/LightOnOCR-2-1B & $0.990$ & $0.357$ & $0.326$ & $0.127$ & $0.640$ & $0.806$ & $0.671$ & $1.207$ & $0.542$ \\
allenai/olmOCR-2-7B-1025-FP8 & $0.999$ & $0.392$ & $0.357$ & $0.127$ & $0.614$ & $0.764$ & $0.438$ & $1.107$ & $0.461$ \\
mistral-ocr-latest & $0.993$ & $0.388$ & $0.286$ & $0.170$ & $0.444$ & $0.733$ & $0.600$ &  & $0.459$ \\
tencent/HunyuanOCR & $0.978$ & $0.251$ & $0.278$ & $0.036$ & $0.372$ & $0.727$ & $0.671$ & $4.467$ & $0.419$ \\
lightonai/LightOnOCR-1B-1025 & $0.996$ & $0.216$ & $0.297$ & $0.012$ & $0.406$ & $0.673$ & $0.602$ & $1.085$ & $0.418$ \\
gemini-2.5-flash-lite & $0.970$ & $0.388$ & $0.422$ & $0.127$ & $0.205$ & $0.588$ & $0.589$ &  & $0.397$ \\
PaddleOCR-VL-1.5 & $0.961$ & $0.206$ & $0.278$ & $0.012$ & $0.125$ & $0.673$ & $0.714$ & $4.056$ & $0.381$ \\
deepseek-ai/DeepSeek-OCR & $\textbf{1.000}$ & $0.124$ & $0.368$ & $0.012$ & $0.382$ & $0.630$ & $0.506$ & $\textbf{0.893}$ & $0.377$ \\
deepseek-ai/DeepSeek-OCR-2 & $0.991$ & $0.096$ & $0.278$ & $0.000$ & $0.281$ & $0.382$ & $0.284$ & $1.470$ & $0.242$ \\
opendatalab/MinerU2.5-2509-1.2B & $0.795$ & $0.100$ & $0.246$ & $0.000$ & $0.093$ & $0.236$ & $0.405$ & $0.898$ & $0.222$ \\
ibm-granite/granite-docling-258M & $0.877$ & $0.031$ & $0.187$ & $0.006$ & $0.067$ & $0.333$ & $0.180$ & $1.197$ & $0.128$ \\
docling & $0.999$ & $0.031$ & $0.119$ & $0.000$ & $0.195$ & $0.055$ & $0.138$ & $3.359$ & $0.119$ \\
\bottomrule
\end{tabular}
}
\end{table}

\subsection{Category-specific analysis}
\begin{figure}[t]
\centering
\includegraphics[width=\linewidth]{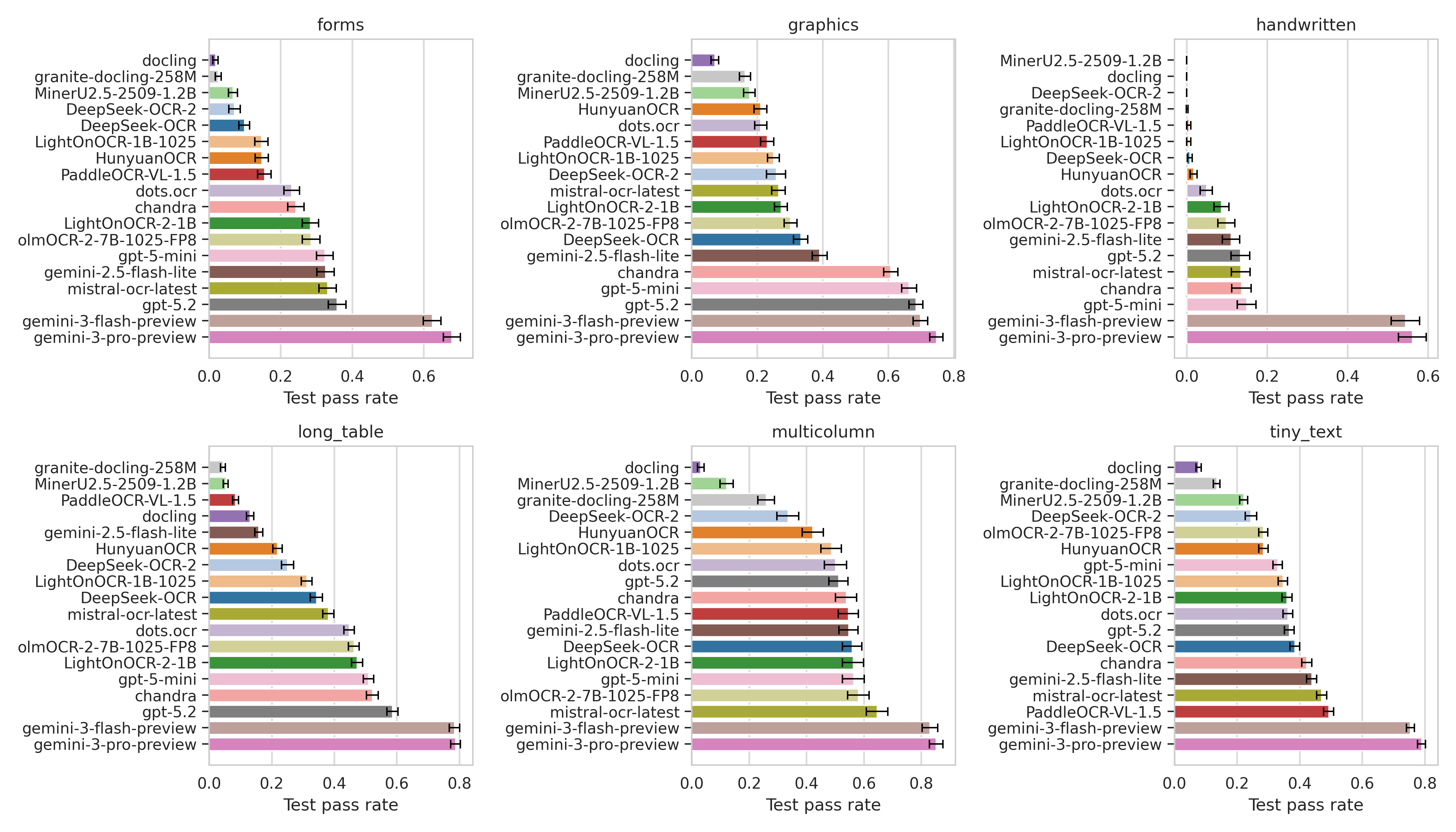}
\caption{Per-category performance by model (mean scores).}
\label{fig:category_results_by_model}
\end{figure}

Performance varied drastically across document categories. Handwritten and forms proved most discriminatory. Gemini 3 Pro retained relative competence (0.60 on handwritten, 0.72 on forms), whereas many smaller models failed completely. For example, granite-docling and MinerU2.5 scored near zero on unstructured handwritten text. As expected graphics were not extracted by non-generalist VLMs with the exception of chandra. Multi-column and tables remained challenging but were better solved than handwriting: high-performing models achieved scores above 0.80 on multi-column text, indicating robust reading order detection.
Figure~\ref{fig:category_results_by_model} provides a category-by-model breakdown.

\subsection{Throughput and resolution effects}
Inference speed varies inversely with model size and architectural complexity. granite-docling and MinerU2.5 were among the fastest, averaging under 0.9 seconds per page. Chandra, despite high accuracy, was the slowest at roughly 4.3 seconds per page, likely due to a heavy visual encoder.

\begin{figure}[t]
\centering
\includegraphics[width=0.60\linewidth]{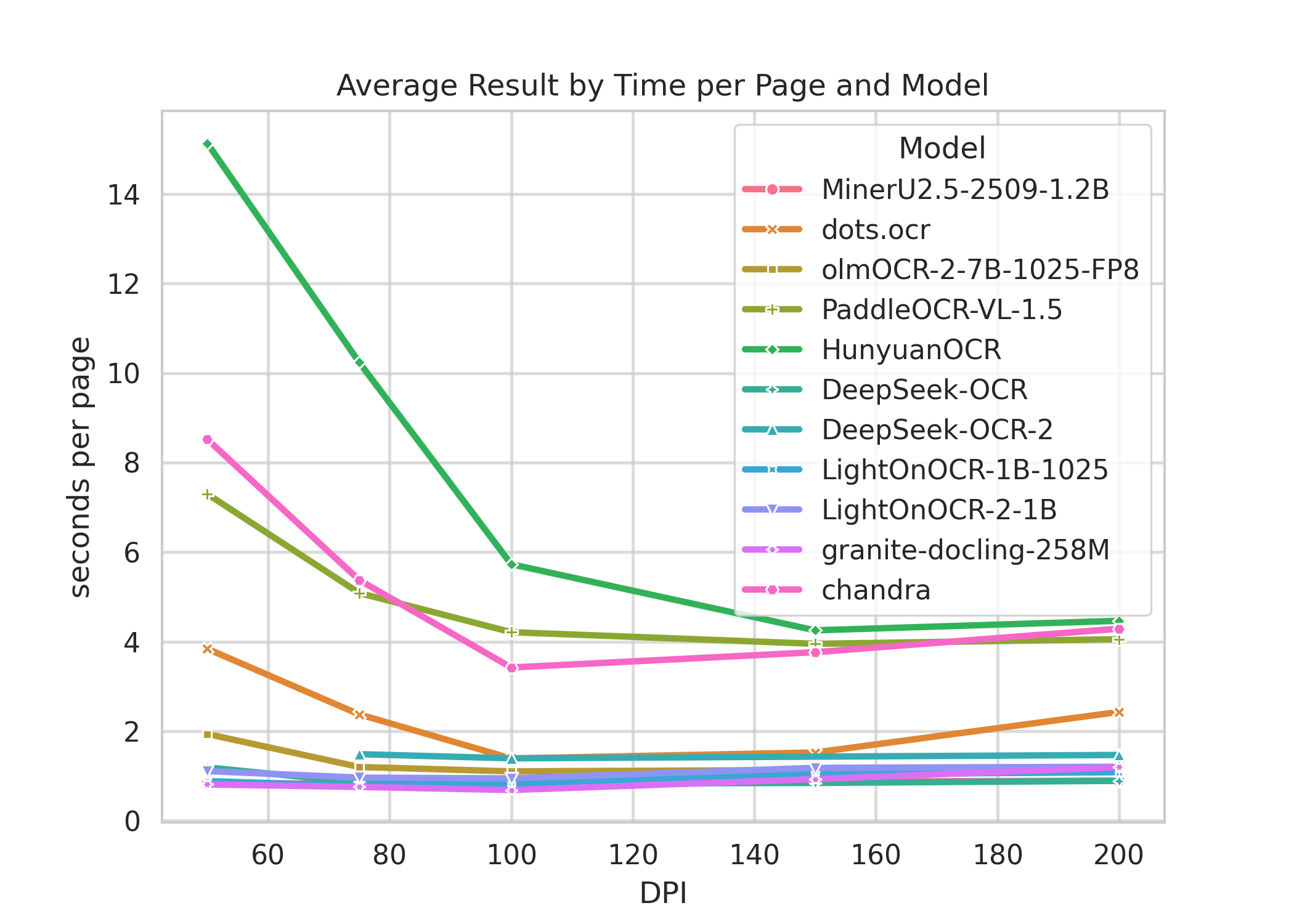}
\caption{DPI sensitivity: time per page as a function of input DPI.}
\label{fig:dpi_effects}
\end{figure}

An analysis of DPI settings revealed a non-monotonic trend for throughput: processing speed initially decreases as image resolution increases up to 100 DPI. This is caused by an increased decoding instability for smaller models resulting in repetitive generations, while higher DPI improves legibility and yields more stable outputs. Figure~\ref{fig:dpi_effects} summarizes throughput trends across DPI.

Additional analyses of decoding choices and inference behavior are provided in the Supplementary Material (Figures~\ref{fig:temperature_vs_performance} and \ref{fig:reasoning_effort_vs_performance_by_category}).

\begin{figure}[t]
\centering
\includegraphics[width=0.60\linewidth]{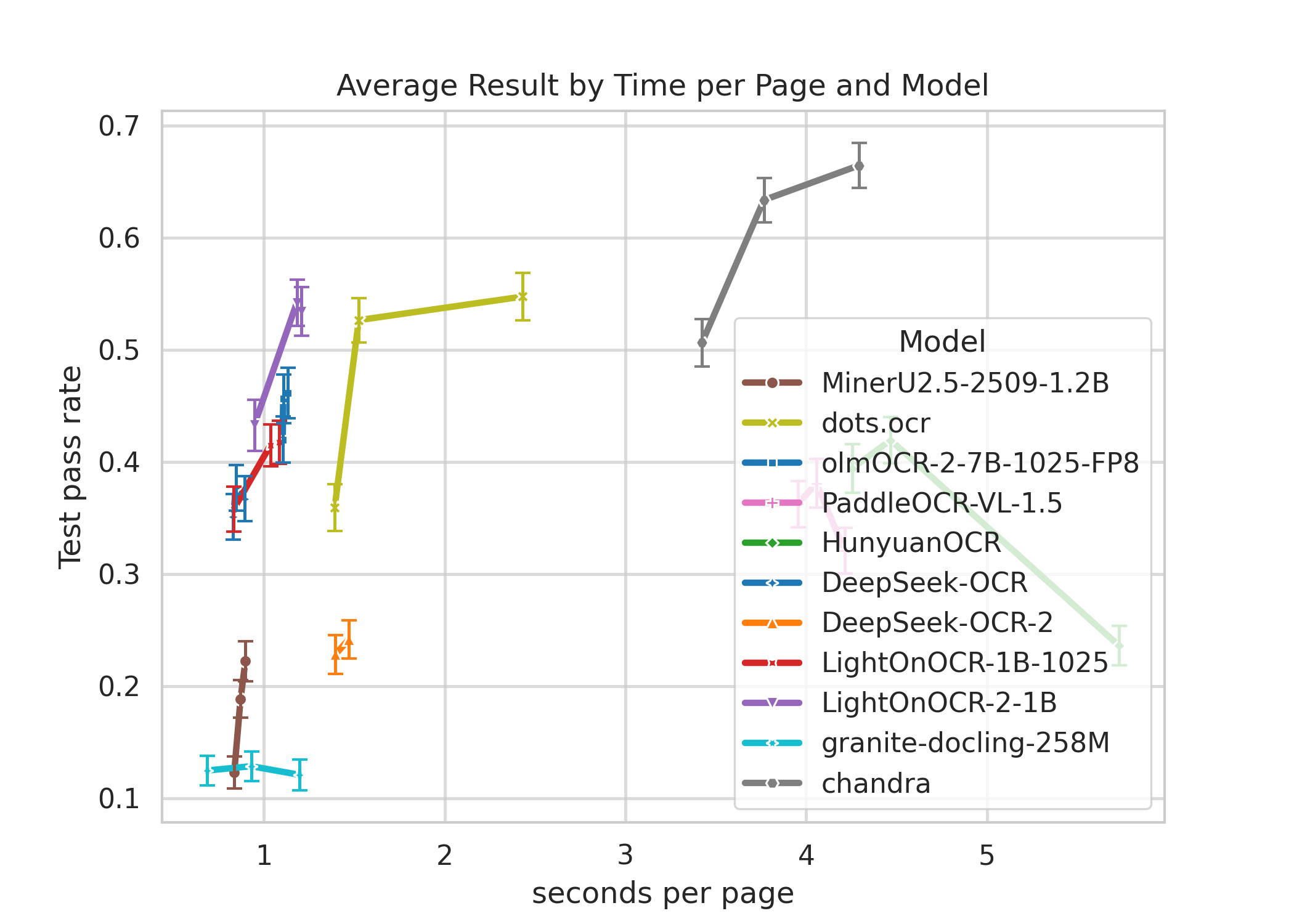}
\caption{Accuracy--throughput trade-off across models.}
\label{fig:avg_vs_time_per_page}
\end{figure}

\section{Discussion}
On this benchmark and under our unified conversion pipeline, performance differences are concentrated in a small set of operationally relevant failure modes. Handwriting and forms remain the most challenging categories, where many models fail in reading hard to decode old documents. For long and dense pages, several smaller VLMs exhibit decoding instability, including repetition/looping behaviors and incomplete coverage of the page, which reduces unit-test pass rates despite otherwise plausible local OCR.

In addition, we observed a systematic failure mode for dots.ocr (see Supplementary Material): incomplete page transcription. A plausible explanation is that the model has been trained with a strong bias to ignore pictures/illustrations; however, in real-world PDFs, ``pictures'' can be text-dense (e.g., scanned callouts, legends embedded in figures, or rasterized colored blocks). On colorful layouts, this bias can cause dots.ocr to skip entire regions that actually contain relevant printed text, leading to partial Markdown outputs even when the locally transcribed regions look correct.

Layout-heavy categories (multi-column pages and long tables) primarily expose reading-order and structure-preservation errors. Even when text recognition is strong, small deviations in region traversal or table linearization can break local structural constraints. This motivates evaluation with targeted tests and normalization that discounts presentation-only variance while remaining sensitive to missing or corrupted content.

Decoding temperature further modulates the stability--accuracy trade-off (Figure~\ref{fig:temperature_vs_performance}). In our experiments, smaller open models are more sensitive to increased randomness, which can amplify local OCR errors into repeated spans, omissions, or structurally inconsistent Markdown. Conversely, overly low temperatures can increase the risk of ``stuck'' continuations on long pages once an erroneous trajectory is selected.

\section{Limitations}
This benchmark and report have several limitations. First, the exclusive focus on French documents limits generalizability to other linguistic contexts, especially non-Latin scripts. Second, throughput comparisons are system-dependent and may be sensitive to quantization choices and hardware compatibility (e.g., fp8 variants on A100). Third, the PDFs are sourced from CCPDF, itself a subset of Common Crawl, which makes it difficult to assess potential pre-training overlap. While the selection procedure emphasizes pages that remain difficult even for the strongest model in our pool (Gemini 3 Pro), contamination cannot be ruled out. In contrast, the forms category is built from documents written manually by a diverse set of participants and is therefore less exposed to crawl-based overlap. The alignment between conclusions on forms and other categories suggest a limited impact of contamination. We also aknowledge that the exact performance of specific models may depend on details of the inference implementation, we welcome any correction or improvements in our inference library. We will gladly update the benchmark leaderboard in such case.

\section{Conclusion}
This report establishes a baseline for VLM-based document parsing on challenging French documents. Current state-of-the-art proprietary models offer the most robust solution for complex layouts and handwritten content. Open-weights models are viable for standard layouts but struggle with visual noise and handwriting. Future work should investigate fine-tuning smaller VLMs on this taxonomy of difficult documents to bridge the performance gap.

Beyond these initial results, we intend this benchmark to be a living, community-driven resource rather than a one-off leaderboard. We encourage practitioners and researchers to take it over by contributing new pages and categories, adding tests for concrete failure modes, refining normalization, and reporting reproducible results for new model releases.

\section*{Acknowledgements}
This work was carried out by members of Probayes and OpenValue, two subsidiaries of La Poste.

This benchmark used documents provided by Gallica under a restricted use:
\begin{quote}
La réutilisation non commerciale des documents de la Bibliothèque nationale de France est libre et gratuite dans le respect de la mention de source : « Bibliothèque nationale de France ou BnF ». La réutilisation commerciale de ces contenus est payante et fait l'objet d'une licence. Est entendue par réutilisation commerciale la revente de contenus sous forme de produits élaborés ou de fourniture de service. Les chercheurs sont exonérés de toute redevance dans le cadre de leurs publications à caractère scientifique et académique. Pour toute réutilisation commerciale ou question sur les conditions d’utilisation des documents de la BnF, merci de contacter : utilisation.commerciale@bnf.fr
\end{quote}

We also gratefully acknowledge the AllenAI OLMo OCR benchmark / OlmOCR-bench, whose public codebase, test design, and normalization logic we adapted extensively for the French PDF-to-Markdown setting \citep{poznanski_OlmOCR_2025,poznanski_OlmOCR_2025-1}.

\clearpage
\bibliographystyle{unsrtnat}
\bibliography{references}  %%% Uncomment this line and comment out the ``thebibliography'' section below to use the external .bib file (using bibtex) .

\clearpage
\section*{Supplementary Material}

\subsection*{Decoding temperature}
\begin{figure}[H]
\centering
\includegraphics[width=0.45\linewidth]{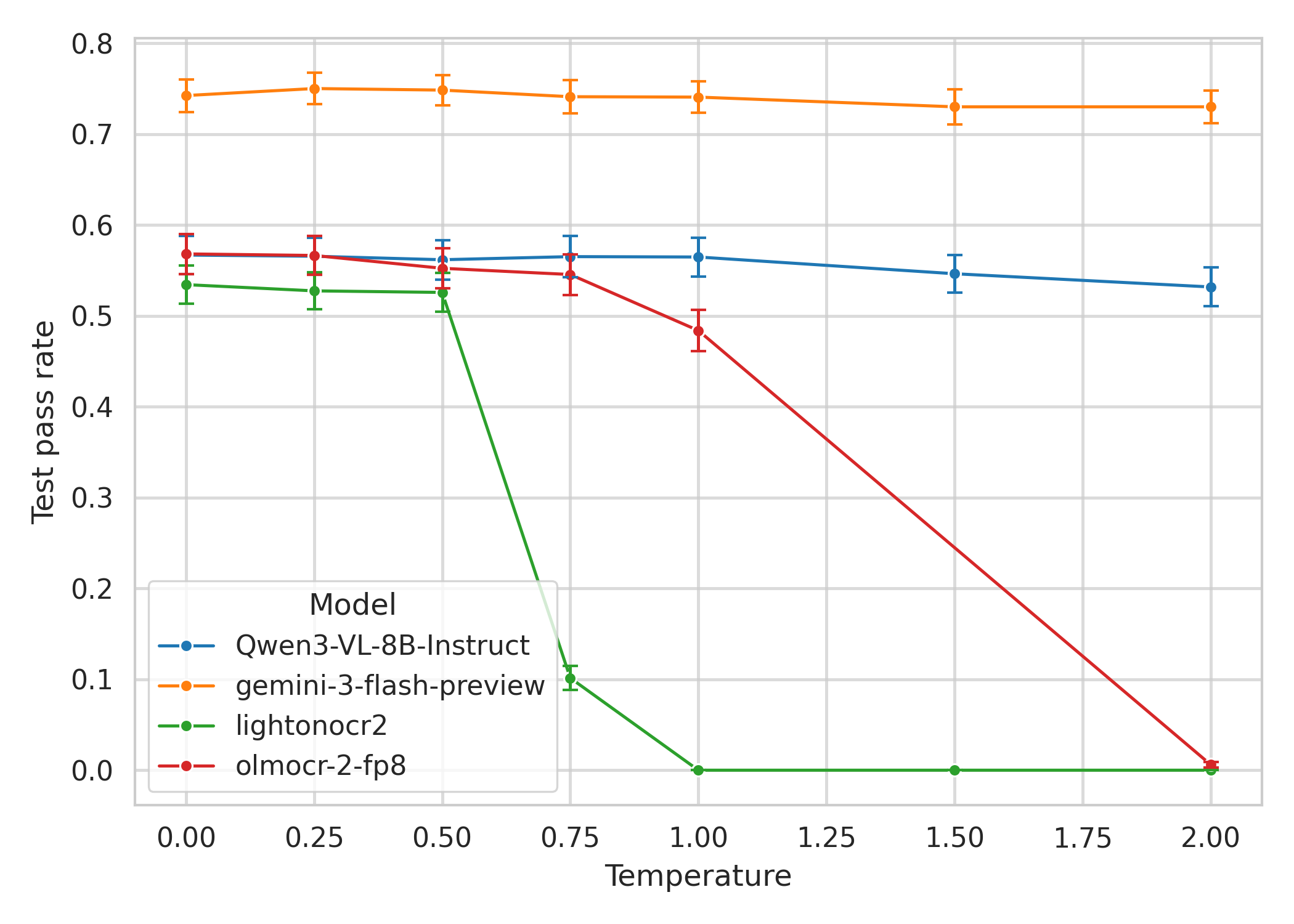}
\caption{Sensitivity of performance to decoding temperature.}
\label{fig:temperature_vs_performance}
\end{figure}

Figure~\ref{fig:temperature_vs_performance} shows that decoding temperature has a measurable effect on end-to-end PDF-to-Markdown quality for small fine-tuned VLMs, where increases in temperature lead to degradation of performance down to zero. However, larger models appear more robust to temperature changes, with only minor performance variations across the tested range.
\subsection*{Reasoning effort}
\begin{figure}[H]
\centering
\includegraphics[width=\linewidth]{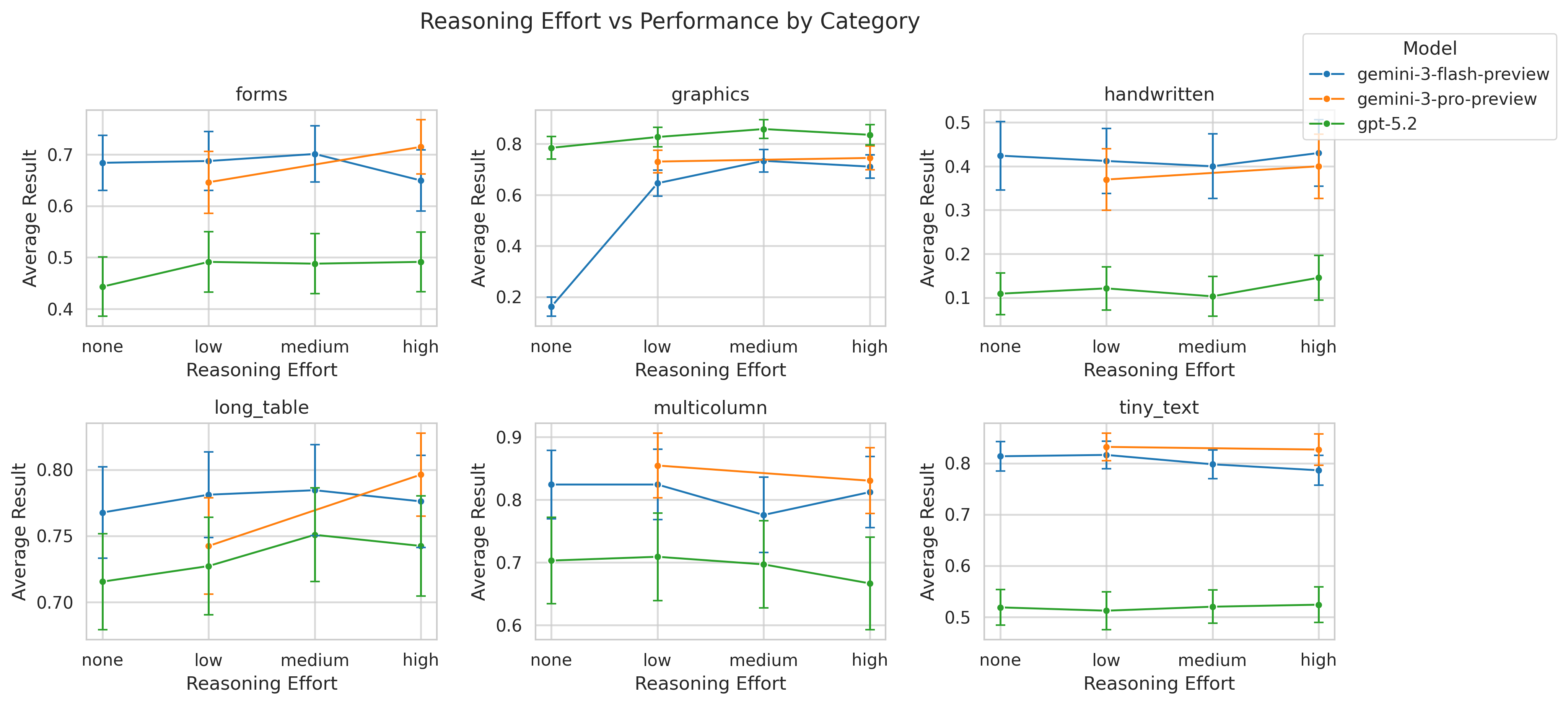}
\caption{Effect of reasoning effort on performance by category.}
\label{fig:reasoning_effort_vs_performance_by_category}
\end{figure}

Figure~\ref{fig:reasoning_effort_vs_performance_by_category} suggests that allocating more reasoning effort has a minor impact on OCR quality across most categories except for graphics and gemini flash models, where a noticeable improvement is observed possibly because the model can better interpret the prompt.

\subsection*{Benchmark reliability and saturation analysis}
To assess the reliability of the evaluation signal and the headroom for future model improvements, we conducted a manual audit of failure cases for both \texttt{OlmOCR-bench} and our proposed benchmark. The objective was to attribute the primary cause of each unit-test failure to either the \emph{model} (e.g., hallucinations, omissions, transcription errors), the \emph{benchmark} (e.g., incorrect ground truth), or \emph{ambiguity} (e.g., valid alternative interpretations of layout or formatting).

For \texttt{OlmOCR-bench}, an audit of 59 failure cases revealed that 47\% were attributable to model errors. A substantial proportion of false negatives stemmed from evaluation artifacts: 31\% were identified as benchmark/annotation issues and 22\% as ambiguous cases (often related to debatable table structures or valid alternative linearizations). The most frequent review labels were \texttt{error\_annotation} (18), \texttt{no table} (13), and \texttt{error word} (8). This suggests that for high-performing models, \texttt{OlmOCR-bench} may be approaching a saturation point where the error signal is dominated by label noise (53\% combined benchmark and ambiguity share) rather than model capability.

Conversely, an audit of 62 failure cases in our proposed benchmark indicated that 98\% of failures were genuine model errors (mostly missing paragraphs, character errors, or word substitutions), with only 2\% attributed to benchmark artifacts. The most frequent review labels were \texttt{missing\_paragraph} (18), \texttt{Error character} (15), and \texttt{Error word} (14). This largely validates the aggressive normalisation, which successfully concentrates the evaluation on distinct failures rather than borderline formatting variances.

\subsubsection*{OlmOCR-bench Failure Audit}
\begin{small}
\begingroup
\catcode`\_=12
\begin{longtable}{llll}
\toprule
type & test_id & review & responsible \\
\midrule
\endfirsthead
\toprule
type & test_id & review & responsible \\
\midrule
\endhead
\midrule
\multicolumn{4}{r}{Continued on next page} \\
\midrule
\endfoot
\bottomrule
\endlastfoot
old_scans & 87_131035...87_131035 & error word & model \\
old_scans & 55_150814...55_150814 & error word & model \\
old_scans & 89_95965...89_95965 & unreadable & ambiguous annotation \\
old_scans & 48_390612...48_390612 & error word & model \\
old_scans & 13_775179...13_775179 & error character & ambiguous annotation \\
old_scans & 62_627024...62_627024 & missing_paragraph & model \\
old_scans & 76_514740...76_514740 & error character & model \\
old_scans & 57_453692...57_453692 & error word & model \\
old_scans & 43_286183...43_286183 & error character & ambiguous annotation \\
old_scans & 17_115907...17_115907 & missing_paragraph & model \\
old_scans & 49_931406...49_931406 & error_annotation & model \\
old_scans & 20_392554...20_392554 & error_annotation & model \\
old_scans & 17_pg17_pg...g1_text_03 & error word & model \\
old_scans & 17_pg24_pg...g1_text_06 & missing_paragraph & model \\
old_scans & 17_pg34_pg...g1_text_07 & missing_paragraph & model \\
long_tiny_text & 13_pg162_p...g1_text_02 & end of line dash & ambiguous annotation \\
long_tiny_text & 20_pg32_pg...g1_text_05 & error_annotation & benchmark \\
long_tiny_text & 13_pg857_p...g1_text_01 & error_annotation & benchmark \\
long_tiny_text & 16a_pg1_te...g1_text_03 & error_annotation & benchmark \\
long_tiny_text & 11_pg186_p...g1_text_02 & error_case & ambiguous annotation \\
long_tiny_text & 17_pg2_pg1...g1_text_17 & error character & ambiguous annotation \\
long_tiny_text & 17_pg34_pg...g1_text_00 & error character & model \\
long_tiny_text & 13_pg857_p...g1_text_01 & error_annotation & benchmark \\
long_tiny_text & 16b_pg1_te...g1_text_02 & error_annotation & benchmark \\
long_tiny_text & 11_pg146_p...g1_text_20 & erreur_encoding & ambiguous annotation \\
long_tiny_text & 11_pg418_p...g1_text_01 & error_annotation & benchmark \\
long_tiny_text & 16a_pg1_te...g1_text_01 & error_annotation & benchmark \\
long_tiny_text & 16a_pg1_te...g1_text_02 & error_annotation & benchmark \\
long_tiny_text & 17_pg33_pg...g1_text_05 & error character & model \\
long_tiny_text & 16a_pg1_te...g1_text_03 & error_annotation & benchmark \\
multi_column & 0b045ab838...r_aa3b8c7b & error_annotation & benchmark \\
multi_column & 0904c70713...r_22ca6197 & error_annotation & benchmark \\
multi_column & 0d7f2cbc54...r_226858b7 & wrong order & model \\
multi_column & 06df5e530c...r_59dca065 & error_annotation & benchmark \\
multi_column & 0621a00904...r_ed9c82c9 & error_case & ambiguous annotation \\
multi_column & 027c390510...r_29bf4787 & error_annotation & benchmark \\
multi_column & 0621a00904...r_e0405e92 & error_annotation & benchmark \\
multi_column & 07f4d706dd...r_e4818f1a & error_annotation & benchmark \\
table & c8c858692b...1_table_13 & no table & model \\
table & 8bc04603d7...1_table_04 & wrong top heading & model \\
table & d46c578922...5_table_00 & no table & model \\
table & 9e3b179dbc...2_table_06 & error word & benchmark \\
table & 8b02fcbf9b...3_table_03 & error_encoding & ambiguous annotation \\
table & 8ec7ac2f20...1_table_08 & wrong top heading & ambiguous annotation \\
table & 53d3f304da...3_table_02 & error word & ambiguous annotation \\
table & 5fd60eb18f...5_table_00 & no table & model \\
table & 1c42d782fc...1_table_03 & no table & model \\
table & d55874d908...4_table_01 & error word & benchmark \\
table & cefac431f0...9_table_00 & no table & model \\
table & 9c38972000...7_table_03 & no table & model \\
table & 3331dcc1ca...2_table_01 & error_annotation & benchmark \\
table & 8bf7270c79...0_table_00 & no table & model \\
table & 0953927d7a...0_table_07 & no table & model \\
table & b5d9db350b...2_table_03 & no table & ambiguous annotation \\
table & 2d0e0586cd...3_table_04 & no table & model \\
table & 8160caa0f0...2_table_01 & no table & model \\
table & 587a1f5a89...7_table_03 & no table & model \\
table & 524a695cbe...1_table_00 & no table & model \\
\end{longtable}

\endgroup
\end{small}

\subsubsection*{French Benchmark Failure Audit}
\begin{small}
\begingroup
\catcode`\_=12
\begin{longtable}{llll}
\toprule
type & test_id & review & responsible \\
\midrule
\endfirsthead
\toprule
type & test_id & review & responsible \\
\midrule
\endhead
\midrule
\multicolumn{4}{r}{Continued on next page} \\
\midrule
\endfoot
\bottomrule
\endlastfoot
multicolumn & 0429f0d4_s...ticolumn_0 & parsed_only_title & model \\
multicolumn & 0429f0d4_s...ticolumn_1 & parsed_only_title & model \\
multicolumn & 0429f0d4_s...ticolumn_2 & parsed_only_title & model \\
multicolumn & 09da7459_I...ticolumn_0 & missing_paragraph & model \\
multicolumn & 110b59ed_f...ticolumn_0 & broken paragraph & model \\
multicolumn & 110b59ed_f...ticolumn_2 & broken paragraph & model \\
multicolumn & 110b59ed_f...ticolumn_3 & Error character & model \\
multicolumn & 110b59ed_f...ticolumn_5 & broken paragraph & model \\
multicolumn & 110b59ed_f...ticolumn_6 & broken paragraph & model \\
multicolumn & 17718f95_P...ticolumn_0 & Error character & model \\
multicolumn & 1d790892_L...ticolumn_0 & blank page & model \\
tiny_text & 00899de2_F...f_p0_tiny1 & Error character & model \\
tiny_text & 00899de2_F...f_p0_tiny2 & Error word & model \\
tiny_text & 00899de2_F...f_p0_tiny0 & Error character & model \\
tiny_text & 00899de2_F...f_p0_tiny0 & Error character & model \\
tiny_text & 028c1fe0_A...f_p0_tiny1 & missing_paragraph & model \\
tiny_text & 02d35b7f_2...f_p0_tiny1 & Error word & model \\
tiny_text & 0423e3c2_p...f_p0_tiny0 & Error word & model \\
tiny_text & 05213325_R...f_p0_tiny0 & Error character & model \\
tiny_text & 05213325_R...f_p0_tiny1 & Error character & model \\
handwritten & btv1b53026...dwritten_0 & Error word & model \\
handwritten & btv1b53026...dwritten_1 & Error character & model \\
handwritten & 021461a6_S...f_p0_tiny0 & missing_paragraph & model \\
handwritten & 021461a6_S...f_p0_tiny1 & Error word & model \\
handwritten & 021461a6_S...f_p0_tiny2 & Error word & model \\
handwritten & 021461a6_S...f_p0_tiny0 & missing_paragraph & model \\
handwritten & 021461a6_S...f_p0_tiny2 & Error word & model \\
handwritten & 021461a6_S...f_p0_tiny0 & Error word & model \\
handwritten & 021461a6_S...f_p0_tiny1 & Error word & model \\
handwritten & 021461a6_S...f_p0_tiny2 & Error word & model \\
forms & Scan_11122...f_p0_text0 & missing_paragraph & model \\
forms & Scan_11122...f_p0_text1 & missing_paragraph & model \\
forms & Scan_11122...f_p0_text2 & missing_paragraph & model \\
forms & Scan_11122...f_p1_text0 & Error character & model \\
forms & Scan_11122...f_p1_text2 & Error word & model \\
forms & Scan_11122...f_p1_text3 & Error character & model \\
forms & Scan_11122...f_p2_text0 & Error word & model \\
forms & Scan_11122...f_p2_text1 & missing_paragraph & model \\
forms & Scan_11122...f_p2_text2 & Error word & model \\
forms & Scan_11122...f_p2_text3 & missing_paragraph & model \\
forms & Scan_11122...f_p1_text4 & erreur_annot & Benchmark \\
table & 025ecab0_A....pdf_p0_t3 & Error character & model \\
table & 036c761a_l....pdf_p0_t0 & Error character & model \\
table & 036c761a_l....pdf_p0_t3 & Error character & model \\
table & 05160169_R....pdf_p0_t1 & Structure error & model \\
table & 07e6f8e3_T....pdf_p0_t1 & Error character & model \\
table & 0d911403_2....pdf_p0_t1 & Error word & model \\
table & 122958aa_G....pdf_p0_t0 & No table & model \\
table & 12ac87e6_F....pdf_p0_t2 & Error character & model \\
table & 12ac87e6_F....pdf_p0_t0 & Top heading not well classified & model \\
table & 12ac87e6_F....pdf_p0_t1 & Top heading not well classified & model \\
table & 175bfc79_8....pdf_p0_t2 & No table & model \\
table & 20b80864_P....pdf_p0_t0 & No table & model \\
graphics & Bilan prem...f_p0_tiny0 & missing_paragraph & model \\
graphics & Bilan prem...f_p0_tiny1 & missing_paragraph & model \\
graphics & Bilan prem...f_p0_tiny2 & missing_paragraph & model \\
graphics & Bilan prem...f_p0_tiny1 & missing_paragraph & model \\
graphics & Bilan prem...f_p0_tiny2 & missing_paragraph & model \\
graphics & PhD_Thesis...f_p0_tiny1 & missing_paragraph & model \\
graphics & PhD_Thesis...f_p0_tiny2 & missing_paragraph & model \\
graphics & PhD_Thesis...f_p0_tiny0 & missing_paragraph & model \\
graphics & PhD_Thesis...f_p0_tiny1 & missing_paragraph & model \\
\end{longtable}

\endgroup
\end{small}

\end{document}